\PassOptionsToClass{nonacm,balance=false}{acmart}
\AtBeginDocument{%
  \settopmatter{printfolios=true}%
}
\documentclass[sigconf]{acmart}

\renewcommand\footnotetextcopyrightpermission[1]{}
\acmConference[SeT-LLM '26]{Secure and Trustworthy Large Language Models}{August 10, 2026}{Jeju, Republic of Korea}
\acmYear{2026}
\acmDOI{}
\acmISBN{}

\title{Risk Is Not Review Value: Wrong-Answer Exposure Under Bounded Review Budgets}

\author{SangJin Park}
\affiliation{%
  \institution{Tynapse}
  \city{Seoul}
  \country{Republic of Korea}
}
\email{sangjin@tynapse.com}

\author{Myungsub Choi}
\affiliation{%
  \institution{Tynapse}
  \city{Seoul}
  \country{Republic of Korea}
}
\email{myungsub@tynapse.com}

\author{Jineok Kim}
\affiliation{%
  \institution{Tynapse}
  \city{Seoul}
  \country{Republic of Korea}
}
\email{gin@tynapse.com}

\author{Minseung Kang}
\affiliation{%
  \institution{Tynapse}
  \city{Seoul}
  \country{Republic of Korea}
}
\email{minseung@tynapse.com}

\begin{document}

\begin{abstract}
LLM assistants often produce more answers than humans can review before users see them.
Most evaluations ask whether an answer is wrong, unsupported, or low-confidence.
Bounded review budgets instead ask which answers should be checked first under a fixed review budget.
Risk alone is not enough: a high-risk answer may be hard to repair, while a moderately risky answer may be directly correctable from available evidence.
For generated-answer evaluation, we model review prioritization as exposure reduction, where review value combines estimated wrongness, intervention affordance, impact, and cost.
We evaluate review queues with Wrong-Answer Exposure Ratio (WAER), the fraction of wrong answers left unreviewed, and post-repair residual exposure (PRRE), the fraction still exposed after deterministic benchmark-supported repairs.
PRRE uses repairability rules that do not numerically reuse the affordance scores used for ranking.
On a 720-item TAT-QA/SciFact stress benchmark, review-value ranking keeps answer-level WAER nearly unchanged at 20\% budget (0.605 vs.\ 0.600) but lowers PRRE from 0.881 to 0.716.
These results show that trustworthy LLM evaluation should measure not only error detection, but also how limited review capacity reduces exposed wrong answers.
\end{abstract}


\maketitle

\section{Introduction}

LLM assistants answer questions over documents, tables, policies, and evidence snippets.
In many realistic workflows, humans cannot review every answer before it reaches a user.
The operational question is therefore not only whether an answer is correct on average, but which answers should consume a bounded review budget.
An answer below the review cutoff remains exposed.

Risk estimates are useful but incomplete for this queueing decision.
Consider two wrong answers: one is highly suspicious but requires reconstructing an ambiguous scope mismatch, while the other is moderately suspicious but contains a numeric value that can be corrected directly from the cited table.
A risk-only queue may spend capacity on the first answer, even though the second has higher expected exposure reduction under a bounded review budget.

This paper studies wrong-answer exposure under bounded review budgets.
Under these budgets, risk is not review value: review prioritization should optimize expected exposure reduction.

Our goal is not to introduce a new verifier model.
Instead, we propose an evaluation objective and a diagnostic benchmark for comparing review-prioritization strategies under the same bounded review budget.

Our contributions are:
\begin{itemize}
    \item We formulate evaluation for already-generated answers under bounded review budgets as exposure-reduction allocation, where review value depends on wrongness, repair affordance, impact, and cost.
    \item We formalize WAER as residual count exposure for truncated review queues and introduce PRRE, whose outcome-side repairability rules do not reuse ranker priors.
    \item We build a public-data stress benchmark with controlled wrong answers and matched controls, then show that review-value ranking improves repair-aware post-review outcomes across budgets and base rates.
\end{itemize}

The proposed setting differs from building a better detector.
A detector can improve $P(\mathrm{wrong})$, but the queueing decision still has to decide which detected risks should consume the next unit of review capacity.
We evaluate residual exposure after the queue is truncated, rather than only score quality before truncation.

\section{Bounded Review Budget Objective}

We first formalize the review queue objective independently of any particular verifier.
Let $A=\{a_1,\ldots,a_n\}$ be generated answers.
Each answer has a gold wrongness label $y_i \in \{0,1\}$ for evaluation, where $y_i=1$ denotes a wrong answer.
A prioritization strategy $\pi$ ranks answers for review.
Given count budget $B$, only the top $B$ answers are reviewed; wrong answers outside this set remain exposed.

Risk-only strategies rank by a wrongness-risk score $r_i\in[0,1]$; when calibrated, $r_i$ may be interpreted as $P(y_i=1\mid a_i,\mathrm{context})$.
This is insufficient when review capacity is scarce.
We instead define the following benchmark review-value scoring rule for answer $i$:

\begin{equation}
  \mathrm{RV}_i =
  \frac{
    r_i \cdot c_i \cdot h_i
  }{
    q_i
  },
\end{equation}

where $c_i$ is bounded intervention affordance, $h_i$ is impact or severity, and $q_i$ is review cost.
An expected-utility interpretation additionally requires calibrated $r_i$, cardinally meaningful $c_i$ and $h_i$, and additive losses; the benchmark does not claim those conditions have been established in deployment.
In the current diagnostic benchmark, $h_i$ is an error-type impact prior and $q_i=1$ for all answers.
With heterogeneous indivisible costs, sorting by $r_i c_i h_i/q_i$ is a greedy density heuristic; exact budgeted selection is a 0--1 knapsack problem.
Risk-only ranking is the special case that uses $r_i$ while treating $c_i$, $h_i$, and $q_i$ as constant.

\begin{figure}[t]
\centering
\includegraphics[width=.96\linewidth]{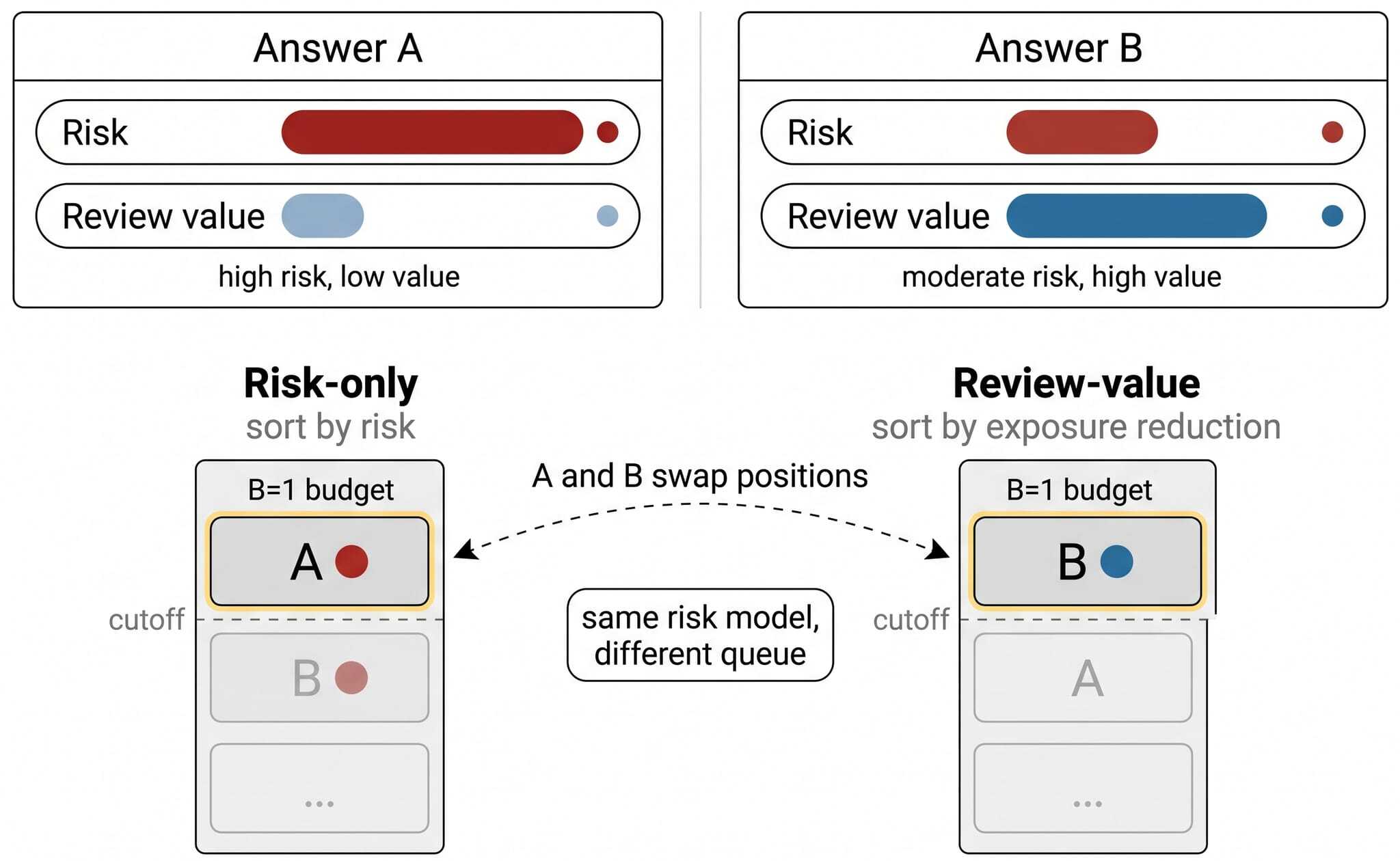}
\Description{Two ranked review queues compare risk-only and review-value selection. The review-value queue promotes a moderately risky but directly repairable answer above a higher-risk answer with low intervention affordance, leaving less repairable exposure below the review cutoff.}
\caption{Pairwise inversion under a bounded review budget.
A risk-only queue selects the answer with the highest estimated wrongness, while a review-value queue can prefer a moderate-risk, high-affordance error over a high-risk, low-affordance one.
With the same risk model and budget, the two objectives can leave different residual exposure below the cutoff.}
\label{fig:review_value_inversion}
\end{figure}

The distinction is visible even without changing the risk model.
For two answers $i$ and $j$, a risk-only policy prefers $i$ whenever $r_i>r_j$.
A review-value policy can prefer $j$ when:

\begin{equation}
  r_i > r_j
  \quad\mathrm{but}\quad
  \frac{r_i c_i h_i}{q_i}
  <
  \frac{r_j c_j h_j}{q_j}.
\end{equation}

Figure~\ref{fig:review_value_inversion} illustrates this inversion as a queueing decision under a fixed review budget.
This pairwise inversion is the benchmark target.
It is not a calibration failure by itself; it is a mismatch between the score being optimized and the review-budget decision.

In this diagnostic benchmark, intervention affordance is an in silico benchmark property, not a human behavior claim.
The empirical outcome is deliberately narrower than general intervention utility: it asks whether public evidence and gold metadata afford a deterministic bounded correction.
We decompose it as:

\begin{equation}
  c_i = g_i \cdot d_i,\quad g_i,d_i \in [0,1],
\end{equation}

where $g_i$ asks whether the intervention target is explicit in public evidence or gold metadata, and $d_i$ asks whether the bounded action is clear and close to unique.
Replacing an incorrect numeric value has high $g_i$ and $d_i$; resolving a scope distortion may require reconstruction and has lower direct intervention affordance.

\section{Exposure Metrics}

We then define outcome metrics for what remains after the review queue is truncated.
For a policy $\pi$ and budget $B$, let $\mathrm{rank}_{\pi}(a_i)$ be the review rank of answer $a_i$.
We define Wrong-Answer Exposure Ratio:

\begin{equation}
  \mathrm{WAER}(\pi,B) =
  \frac{
    \sum_i \mathbb{1}[y_i=1]\mathbb{1}[\mathrm{rank}_{\pi}(a_i)>B]
  }{
    \sum_i \mathbb{1}[y_i=1]
  } .
\end{equation}

WAER measures the fraction of wrong answers left unreviewed.
Lower is better.
Reporting WAER over budgets such as 5, 10, 20, and 40 percent yields an exposure-capacity curve.

WAER treats all wrong answers equally.
To evaluate repair-aware residual exposure without reusing the ranker's numeric affordance values as the outcome, we define \emph{post-repair residual exposure} (PRRE).
A selected wrong answer is removed from exposure only if it is recoverable under benchmark data mechanics; selected-but-unrepairable and unselected wrong answers remain exposed.
PRRE is a correction-side residual exposure metric: it gives exposure credit only for deterministic benchmark-side repair, not for suppression, escalation, or reviewer-specific rejection behavior.
With $\rho_i\in\{0,1\}$ a repairability flag,

\begin{equation}
  \mathrm{PRRE}(\pi,B) =
  1 - \frac{
    \sum_i \mathbb{1}[y_i=1]\,\rho_i\,\mathbb{1}[\mathrm{rank}_{\pi}(a_i)\le B]
  }{
    \sum_i \mathbb{1}[y_i=1]
  } .
\end{equation}

This construction assigns distinct roles to ranking and outcome assessment.
Under a review-value policy, the affordance prior $c_i$ influences the metric through the induced ranking $\mathrm{rank}_{\pi}(a_i)$ and hence the composition of the review set.
Once an answer is selected, however, the outcome-side flag $\rho_i$ alone determines whether a wrong answer receives repair credit.
For example, a numeric error is repairable iff the answer deviates from the gold value present in the evidence, while scope distortion and conclusion mismatch are not treated as deterministically recoverable.
Although $c_i$ and $\rho_i$ are structurally related through controlled error type, they serve distinct roles: $c_i$ encodes type-level actionability for allocation, whereas $\rho_i$ is derived from deterministic data mechanics for outcome assessment.
Accordingly, the benchmark tests the allocation value of type-level actionability rather than independently learned human repairability, while defining repair credit without copying or thresholding $c_i$.

For context, we also report two additional exposure metrics.
Weighted exposure (WDE) replaces the count of residual wrong answers with residual $h_i$-weighted wrong-answer exposure.
Review-value exposure (RVE) weights residual exposure by the same actionable-severity term used by the ranker, $c_i h_i/q_i$, and is therefore contextual rather than the primary recoverability outcome.
PRRE is the main repair-aware outcome; RVE is a contextual severity-weighted outcome tied to the $c_i h_i/q_i$ ranking term.
In the current pilots, $h_i$ is the error-type impact prior in Table~\ref{tab:affordance} and $q_i=1$ for all answers.

\section{Diagnostic Benchmark}

To test the objective-metric distinction under controlled error types, we construct a public-data diagnostic benchmark.
The benchmark is built from public datasets with available evidence or gold labels: TAT-QA for table-and-text financial QA requiring numerical reasoning \cite{zhu2021tatqa}, and SciFact for evidence-backed scientific claim verification \cite{wadden2020scifact}.

For each source item, we generate correct answers, controlled wrong answers, and style-matched correct controls.
Matched controls reduce shortcut artifacts: a strategy should not succeed merely because wrong answers are longer, more assertive, or templated.
Gold wrongness labels come from public labels plus controlled mutation metadata; LLMs may paraphrase answers or score verifier baselines, but they are not gold labelers.
The reported 720 instances comprise 60 TAT-QA facts with eight variants each and 60 SciFact facts with four variants each.

The design follows six requirements: evidence-grounded items, matched correct controls, error-type diversity, budget curves, base-rate sweeps, and leakage tiers separating operational, diagnostic, and oracle baselines.

\begin{table}
\centering
\caption{Benchmark-defined intervention and impact priors.
$g_i$ denotes evidence availability, $d_i$ denotes action determinacy, $c_i=g_id_i$, and $h_i$ is an error-type impact prior.
Current pilots set $q_i=1$.}
\label{tab:affordance}
\begin{tabular}{@{}lcccc@{}}
\toprule
Error type & $g_i$ & $d_i$ & $c_i$ & $h_i$ \\
\midrule
Numeric perturbation & 1.0 & 1.0 & 1.00 & 1 \\
Direction flip & .9 & .9 & .81 & 1 \\
Unsupported addition & .7 & .6 & .42 & 2 \\
Scope distortion & .3 & .3 & .09 & 2 \\
Conclusion mismatch & .5 & .4 & .20 & 3 \\
\bottomrule
\end{tabular}
\end{table}

Controlled error types and their intervention and impact priors are shown in Table~\ref{tab:affordance}.
The low scope-distortion value reflects both weak target availability and weak action determinacy: public evidence may reveal that scope was overstated without specifying a unique bounded correction.
The priors make the allocation assumption explicit and testable; future work should validate or replace them with human-review measurements.
Controlled mutations do not replace naturalistic evaluation, but they let the benchmark separate wrongness, impact, and repairability while preserving public evidence and matched controls.

\section{Prioritization Strategies}

We compare strategies that differ in whether they estimate only wrongness risk or also approximate review value.
Risk-only strategies include random, confidence, rule/surface, and LLM-judge-like rankings; these primarily estimate $r_i$.
Review-value strategies multiply a risk score by estimated intervention affordance and impact.
The operational risk-only and review-value proxies share the same source-consistency risk score $r_i$; the review-value proxy estimates error type from source-consistency cues and multiplies that score by type-level $\hat{c}_i$ and $\hat{h}_i$ before sorting.
The gold-factor variant instead uses gold wrongness and controlled error-type factors in the same scoring rule and is evaluation-only.
Explicit checkers use benchmark-defined evidence slots more directly than operational risk-only baselines and are included as diagnostic assumption probes.
No operational risk-only baseline observes $y_i$, injected type, severity, or affordance labels.

\section{Diagnostic Results}

We separate baselines by what information they observe.
Random/confidence, rule/surface, and strong-verifier scores are operational risk-only baselines.
Explicit checkers and review-value proxies are diagnostic baselines; oracle variants are evaluation-only references.
The explicit checker is not the main operational comparator; its appendix results test whether evidence-slot repairability, rather than wrongness alone, explains the benchmark mechanism.
Table~\ref{tab:review_value} reports the main deterministic stress result: 720 public-data-derived items, 360 wrong answers, 360 matched controls, and 32 outer seeds.
At 20\% review capacity, risk-only and review-value ranking are nearly tied on answer-level WAER (0.605 vs.\ 0.600).
The difference appears in post-review exposure: review-value ranking leaves lower PRRE (0.716 vs.\ 0.881) because it spends capacity on errors that are recoverable from public evidence rather than on high-risk but unrepairable errors.
We treat RVE as a contextual metric because it weights by the same affordance prior the review-value ranker uses.
Here $B=144$ and $W=360$, so the minimum attainable WAER is $1-B/W=.600$; risk-only and review-value are therefore at or near the wrong-answer capture floor.
Their mean PRRE values correspond to selecting approximately 43 versus 102 deterministically repairable wrong answers, isolating queue composition after detector capture.

\begin{table}
\centering
\caption{Public-data stress result at 20\% review capacity, mean over 32 outer seeds.
Brackets show 95\% bootstrap confidence intervals for the 32-seed mean and reflect outer-seed variation only.
Lower is better.}
\label{tab:review_value}
\begin{tabular}{lcc}
\toprule
Strategy & WAER & PRRE \\
\midrule
Random & .800 [.800,.800] & .911 [.911,.912] \\
Risk-only proxy & .605 [.604,.607] & .881 [.877,.885] \\
Review-value proxy & \textbf{.600} [.600,.600] & \textbf{.716} [.713,.719] \\
Gold-factor RV & .600 [.600,.600] & .656 [.654,.659] \\
Repair-count oracle & .600 [.600,.600] & .600 [.600,.600] \\
\bottomrule
\end{tabular}
\end{table}

Across the 32 outer seeds, the mean number of deterministically repairable wrong answers is 159.7; uniform random review therefore has expected PRRE $1-(144/720)(159.7/360)=.911$, matching the observed .911.

Table~\ref{tab:bootstrap_delta} reports paired source-fact uncertainty for the same 20\% budget.
The source-cluster interval excludes zero for the small WAER difference, but its absolute magnitude is only .005 and both queues are at or near the .600 floor; we therefore interpret their wrong-answer capture as operationally near-tied.
PRRE and RVE decrease under review-value ranking.
Appendix Table~\ref{tab:source_cluster_bootstrap} reports the corresponding policy-wise bootstrap point estimates and intervals.

\begin{table}
\centering
\caption{Paired, dataset-stratified source-fact cluster bootstrap of the 32-seed mean at 20\% review capacity.
Policies are reranked within each of 1{,}000 resamples; deltas are review-value minus risk-only, and negative values favor review-value.}
\label{tab:bootstrap_delta}
\begin{tabular}{lc}
\toprule
Metric & $\Delta$ [95\% interval] \\
\midrule
WAER & -.005 [-.007,-.004] \\
WDE & .119 [.114,.122] \\
RVE & -.116 [-.120,-.112] \\
PRRE & -.165 [-.168,-.160] \\
\bottomrule
\end{tabular}
\end{table}

The positive WDE delta is not a contradiction: WDE asks how much severity-weighted wrongness remains, while PRRE asks whether selected errors are recoverable from available evidence.

Table~\ref{tab:same_risk} isolates allocation from risk estimation by holding the verifier risk score $r_i$ fixed.
Ranking by $r_i$ and $r_i c_i$ leaves answer-level exposure unchanged (WAER .600), but the intervention-aware queue lowers PRRE from .767 to .600.
Thus the gain comes from changing the queue allocation objective rather than from a stronger verifier.

\begin{table}
\centering
\caption{Same-risk verifier ablation at 20\% review capacity.
The first three rows share verifier risk $r_i$; Gold-factor RV is an evaluation-only reference.
Lower is better.}
\label{tab:same_risk}
\begin{tabular}{lcc}
\toprule
Queue objective & WAER & PRRE \\
\midrule
Verifier risk $r_i$ & .600 & .767 \\
$r_i c_i$ & .600 & \textbf{.600} \\
$r_i c_i h_i$ & .600 & .667 \\
Gold-factor RV & .600 & .656 \\
\bottomrule
\end{tabular}
\end{table}

The review-value objective prioritizes actionable exposure: errors that are both consequential and amenable to bounded intervention.
Therefore, lower RVE or PRRE can coincide with higher WDE when severe errors have low intervention affordance under the benchmark evidence.
This also explains why an $r_i c_i h_i$ queue need not minimize PRRE relative to an $r_i c_i$ queue: $h_i$ intentionally adds severity to the actionability objective, while PRRE isolates recoverable residual exposure as a benchmark-defined outcome that does not reuse numeric affordance values.

\paragraph{Which errors remain exposed?}
Table~\ref{tab:error_mechanism} shows that the allocation effect is heterogeneous across error types.
At 20\% review capacity, the review-value queue captures nearly all numeric perturbations (WAER .067), while every scope distortion remains below the review cutoff (WAER 1.000).
Direction flips, conclusion mismatches, and unsupported additions fall between these extremes.
Conclusion mismatches are sometimes reviewed (WAER .625) but have no deterministic outcome-side repair rule, so selecting them does not automatically earn repair credit under PRRE.
The aggregate PRRE improvement therefore reflects reallocation toward errors with explicit, deterministic interventions rather than a uniform improvement in error detection.

\begin{table}
\centering
\caption{Base-seed error-type outcomes for the review-value proxy at 20\% budget.
$c_i$ is the ranker affordance prior; $\rho_i/n$ reports actual outcome-side deterministic repairability.
Lower WAER is better.}
\label{tab:error_mechanism}
\begin{tabular}{@{}lccc@{}}
\toprule
Error type & $c_i$ & WAER & $\rho_i/n$ \\
\midrule
Numeric perturbation & 1.00 & .067 & 60/60 \\
Direction flip & .81 & .500 & 60/60 \\
Unsupported addition & .42 & .783 & 32/60 \\
Conclusion mismatch & .20 & .625 & 0/120 \\
Scope distortion & .09 & 1.000 & 0/60 \\
\bottomrule
\end{tabular}
\end{table}

Table~\ref{tab:prre_curve} reports PRRE across budgets.
Across all tested budgets, review-value ranking leaves lower PRRE than the risk-only proxy.
At 40\% capacity, the review-value proxy reaches the repair-count lower bound (.556), suggesting that the proxy covers the benchmark-recoverable errors in this balanced stress setting.
Table~\ref{tab:review_value} reports the corresponding answer-level exposure result at 20\% budget.
Appendices A and B provide the complete deterministic ladder and the corresponding LLM-assisted pilot checks.

\begin{table}
\centering
\caption{PRRE over review budgets, mean over 32 seeds.
Review-value ranking lowers post-repair residual exposure at every budget; reference rows are evaluation-only.
Lower is better.}
\label{tab:prre_curve}
\begin{tabular}{lcccc}
\toprule
Strategy (PRRE) & 5\% & 10\% & 20\% & 40\% \\
\midrule
Risk-only proxy & .989 & .962 & .881 & .704 \\
Review-value proxy & \textbf{.903} & \textbf{.828} & \textbf{.716} & \textbf{.556} \\
Gold-factor RV & .900 & .811 & .656 & .556 \\
Repair-count oracle & .900 & .800 & .600 & .556 \\
\bottomrule
\end{tabular}
\end{table}

Because the balanced stress setting is diagnostic, we also sweep wrong-answer base rates from 5\% to 30\% (Appendix Table~\ref{tab:base_rate}).
WAER is sensitive to wrong-answer base rate and count-budget saturation, so we report it as operational residual count exposure rather than as the sole review-value signal.
PRRE is the more stable diagnostic signal for the review-value objective: it asks whether review-budget allocation changes which recoverable errors remain exposed.
In the sweep, review-value ranking lowers PRRE at every rate while WAER is comparable or better except at the sparsest 5\% setting, where the difference is negligible.
Prior-sensitivity checks perturb $g_i$ and $d_i$ by fixed offsets and random noise (Tables~\ref{tab:sensitivity}--\ref{tab:random_prior_sensitivity}).
Contextual RVE improves under all 1{,}000 random perturbations at each tested radius.
PRRE improvements are strong but not universal under larger prior shifts: review-value wins in 100.0\%, 94.3\%, and 84.3\% of random perturbations at radii $\pm .1$, $\pm .2$, and $\pm .3$, respectively.
The fixed-offset check shows the same caveat, with the largest positive offset weakening the PRRE advantage.
Additional template-cue and heterogeneous-cost sanity checks (Table~\ref{tab:template_cost_sanity}) preserve the qualitative PRRE advantage.

\section{Related Work}

Selective classification and abstention trade coverage against prediction risk \cite{elyaniv2010selective,geifman2019selectivenet}, while selective QA and selective generation study when not to answer \cite{kamath2020selectiveqa,ren2023selfevaluation}.
Calibration methods seek confidence signals aligned with answer correctness \cite{guo2017calibration,jiang2021lmcalibration,ulmer2024calibrating}.
For free-form LLM outputs, semantic, black-box, and claim-level uncertainty methods estimate failure or factuality signals \cite{xiong2024expressuncertainty,lin2024confidence,farquhar2024semanticentropy,fadeeva2024factchecking}.
Human-assistance and algorithmic-triage work studies when model predictions should give way to human judgment \cite{wilder2020complement,de2021assistance,okati2021triage}.
Learning-to-defer further studies consistent, calibrated, and post-hoc routing to a human or expert \cite{madras2018defer,mozannar2020defer,verma2022calibrated,narasimhan2022posthoc}.
Cost-sensitive learning-to-defer adds workload constraints and global classifier/expert assignment; our setting instead evaluates post-generation review policies \cite{alves2024costsensitive}.
Confidence-based check-set selection prioritizes low-confidence LLM outputs under budgeted human effort and reports effective accuracy after checking \cite{delacruz2025checkset}.
Relative to this line, WAER normalizes residual wrong-answer exposure, while PRRE withholds credit without a deterministic outcome-side repair rule; same-risk ablations isolate the queue objective from risk estimation.

A broad hallucination survey organizes the causes and evaluation of factual errors \cite{ji2023hallucinationsurvey}.
Factuality benchmarks and fact-verification datasets measure truthfulness, evidence support, and hallucination risk \cite{lin2022truthfulqa,thorne2018fever,wadden2020scifact,min2023factscore,manakul2023selfcheckgpt}.
Attribution and retrieval-grounded evaluation further test citation support and evidence-grounded generation \cite{rashkin2023attribution,gao2023alce}, while LLM-based NLG evaluation and judging work studies alignment with human judgments and judge reliability \cite{liu2023geval,zheng2023judge}.
Post-generation factual correction can revise model outputs using tools or retrieved evidence \cite{gao2023rarr,gou2024critic,li2025rac}; PRRE instead evaluates a review queue and grants repair credit only under prespecified outcome-side rules.
Active learning allocates scarce human labels to improve a model \cite{settles2009active}.
Review-budget allocation instead evaluates which exposed answers should be inspected so wrong information is not left below the review cutoff.

\section{Discussion and Limitations}

These results are limited to a controlled public-data stress benchmark.
They do not show that risk-only ranking must fail, or that a stronger verifier could not improve the queue.
Instead, the same-risk ablations show that changing the queue objective, while holding $r_i$ fixed, changes which wrong answers remain below the cutoff.

The type priors are benchmark assumptions rather than estimates of operational utility, and their relationship to $\rho_i$ reflects the controlled error-type design.
A verifier score remains useful; the same-risk ablations only isolate the effect of changing the queue objective after $r_i$ is fixed.
We treat the LLM-assisted verifier pilot as auxiliary evidence while keeping the deterministic public-data stress run as the main result.

The benchmark uses synthetic answer variants.
PRRE uses \emph{deterministic} recoverability rules from public data mechanics; it is not a measurement of whether a human or model reviewer would repair each error in practice.
Treating scope and conclusion errors as not deterministically recoverable is a conservative diagnostic choice, not a claim that such errors are never repairable by humans.
Strong verifier baselines may vary by model provider and prompt.
Repair-side model checks are auxiliary stress tests, not human validation or gold-label sources: a 100-answer repair pilot and a six-model panel found that 251/300 judgments rated repairs as reducing exposure and 240/300 rated them as supported by public evidence (Appendix B).
These checks do not change WAER, PRRE, or any gold label.
The benchmark uses only public TAT-QA/SciFact data and controlled answer variants; it contains no customer data, production traces, or human-subject data.
The intended scope is a diagnostic benchmark for separating risk scoring from queue allocation under bounded review budgets, not a production prevalence estimate.

\section{Conclusion}

We propose evaluating LLM outputs under bounded review budgets as an exposure-reduction allocation problem.
Risk is not review value: review priority should depend on wrongness, intervention affordance, impact, and cost.
On a public-data stress benchmark, risk-only and review-value ranking are nearly tied on WAER at 20\% budget, yet review-value ranking leaves lower PRRE across budgets and base rates because the two scores allocate captured errors differently under the benchmark's type-level actionability assumptions.

\begin{acks}
This work was supported by the Korea Association for AI \& ICT Promotion (KAIT) and the National IT Industry Promotion Agency (NIPA) grant funded by the Korea government (MSIT), under the ``Advanced GPU Infrastructure Utilization Support Program'' (Grant No. 04-26-03-0029).
\end{acks}

\bibliographystyle{ACM-Reference-Format}
\bibliography{references}

@inproceedings{geifman2019selectivenet,
  title = {{SelectiveNet}: A Deep Neural Network with an Integrated Reject Option},
  author = {Geifman, Yonatan and El-Yaniv, Ran},
  booktitle = {Proceedings of the 36th International Conference on Machine Learning},
  pages = {2151--2159},
  year = {2019},
  volume = {97},
  series = {Proceedings of Machine Learning Research},
  address = {Long Beach, CA, USA},
  publisher = {PMLR},
  url = {https://proceedings.mlr.press/v97/geifman19a.html}
}

@article{elyaniv2010selective,
  title = {On the Foundations of Noise-free Selective Classification},
  author = {El-Yaniv, Ran and Wiener, Yair},
  journal = {Journal of Machine Learning Research},
  volume = {11},
  number = {53},
  pages = {1605--1641},
  year = {2010},
  url = {https://jmlr.org/papers/v11/el-yaniv10a.html}
}

@inproceedings{guo2017calibration,
  title = {On Calibration of Modern Neural Networks},
  author = {Guo, Chuan and Pleiss, Geoff and Sun, Yu and Weinberger, Kilian Q.},
  booktitle = {Proceedings of the 34th International Conference on Machine Learning},
  pages = {1321--1330},
  year = {2017},
  volume = {70},
  series = {Proceedings of Machine Learning Research},
  address = {Sydney, NSW, Australia},
  publisher = {PMLR},
  url = {https://proceedings.mlr.press/v70/guo17a.html}
}

@techreport{settles2009active,
  title = {Active Learning Literature Survey},
  author = {Settles, Burr},
  institution = {University of Wisconsin-Madison},
  type = {Computer Sciences Technical Report},
  number = {1648},
  address = {Madison, WI, USA},
  year = {2009},
  url = {https://minds.wisconsin.edu/handle/1793/60660}
}

@inproceedings{madras2018defer,
  title = {Predict Responsibly: Improving Fairness and Accuracy by Learning to Defer},
  author = {Madras, David and Pitassi, Toni and Zemel, Richard},
  booktitle = {Advances in Neural Information Processing Systems},
  volume = {31},
  pages = {6150--6160},
  year = {2018},
  address = {Montreal, Canada},
  publisher = {Curran Associates, Inc.},
  url = {https://proceedings.neurips.cc/paper_files/paper/2018/file/09d37c08f7b129e96277388757530c72-Paper.pdf}
}

@inproceedings{mozannar2020defer,
  title = {Consistent Estimators for Learning to Defer to an Expert},
  author = {Mozannar, Hussein and Sontag, David},
  booktitle = {Proceedings of the 37th International Conference on Machine Learning},
  pages = {7076--7087},
  year = {2020},
  volume = {119},
  series = {Proceedings of Machine Learning Research},
  address = {Virtual Event},
  publisher = {PMLR},
  url = {https://proceedings.mlr.press/v119/mozannar20b.html}
}

@article{alves2024costsensitive,
  title = {Cost-Sensitive Learning to Defer to Multiple Experts with Workload Constraints},
  author = {Alves, Jean Vieira and Leit{\~a}o, Diogo and Jesus, S{\'e}rgio and Sampaio, Marco O. P. and Li{\'e}bana, Javier and Saleiro, Pedro and Figueiredo, M{\'a}rio A. T. and Bizarro, Pedro},
  journal = {Transactions on Machine Learning Research},
  year = {2024},
  issn = {2835-8856},
  url = {https://openreview.net/forum?id=TAvGZm2Rqb}
}

@inproceedings{delacruz2025checkset,
  title = {Evaluating Large Language Models for Confidence-based Check Set Selection},
  author = {dela Cruz, Jane Arleth and Hendrickx, Iris and Larson, Martha},
  booktitle = {Findings of the Association for Computational Linguistics: ACL 2025},
  pages = {16249--16265},
  year = {2025},
  address = {Vienna, Austria},
  publisher = {Association for Computational Linguistics},
  doi = {10.18653/v1/2025.findings-acl.836},
  url = {https://aclanthology.org/2025.findings-acl.836/}
}

@inproceedings{kamath2020selectiveqa,
  title = {Selective Question Answering under Domain Shift},
  author = {Kamath, Amita and Jia, Robin and Liang, Percy},
  booktitle = {Proceedings of the 58th Annual Meeting of the Association for Computational Linguistics},
  pages = {5684--5696},
  year = {2020},
  address = {Online},
  publisher = {Association for Computational Linguistics},
  doi = {10.18653/v1/2020.acl-main.503},
  url = {https://aclanthology.org/2020.acl-main.503/}
}

@inproceedings{ren2023selfevaluation,
  title = {Self-Evaluation Improves Selective Generation in Large Language Models},
  author = {Ren, Jie and Zhao, Yao and Vu, Tu and Liu, Peter J. and Lakshminarayanan, Balaji},
  booktitle = {Proceedings on ``I Can't Believe It's Not Better: Failure Modes in the Age of Foundation Models'' at NeurIPS 2023 Workshops},
  pages = {49--64},
  year = {2023},
  volume = {239},
  series = {Proceedings of Machine Learning Research},
  address = {New Orleans, LA, USA},
  publisher = {PMLR},
  url = {https://proceedings.mlr.press/v239/ren23a.html}
}

@inproceedings{ulmer2024calibrating,
  title = {Calibrating Large Language Models Using Their Generations Only},
  author = {Ulmer, Dennis and Gubri, Martin and Lee, Hwaran and Yun, Sangdoo and Oh, Seong},
  booktitle = {Proceedings of the 62nd Annual Meeting of the Association for Computational Linguistics (Volume 1: Long Papers)},
  pages = {15440--15459},
  year = {2024},
  address = {Bangkok, Thailand},
  publisher = {Association for Computational Linguistics},
  doi = {10.18653/v1/2024.acl-long.824},
  url = {https://aclanthology.org/2024.acl-long.824/}
}

@inproceedings{lin2022truthfulqa,
  title = {{TruthfulQA}: Measuring How Models Mimic Human Falsehoods},
  author = {Lin, Stephanie and Hilton, Jacob and Evans, Owain},
  booktitle = {Proceedings of the 60th Annual Meeting of the Association for Computational Linguistics (Volume 1: Long Papers)},
  pages = {3214--3252},
  year = {2022},
  address = {Dublin, Ireland},
  publisher = {Association for Computational Linguistics},
  doi = {10.18653/v1/2022.acl-long.229},
  url = {https://aclanthology.org/2022.acl-long.229/}
}

@inproceedings{thorne2018fever,
  title = {{FEVER}: A Large-scale Dataset for Fact Extraction and {VER}ification},
  author = {Thorne, James and Vlachos, Andreas and Christodoulopoulos, Christos and Mittal, Arpit},
  booktitle = {Proceedings of the 2018 Conference of the North American Chapter of the Association for Computational Linguistics: Human Language Technologies, Volume 1 (Long Papers)},
  pages = {809--819},
  year = {2018},
  address = {New Orleans, LA, USA},
  publisher = {Association for Computational Linguistics},
  doi = {10.18653/v1/N18-1074},
  url = {https://aclanthology.org/N18-1074/}
}

@inproceedings{wadden2020scifact,
  title = {Fact or Fiction: Verifying Scientific Claims},
  author = {Wadden, David and Lin, Shanchuan and Lo, Kyle and Wang, Lucy Lu and van Zuylen, Madeleine and Cohan, Arman and Hajishirzi, Hannaneh},
  booktitle = {Proceedings of the 2020 Conference on Empirical Methods in Natural Language Processing (EMNLP)},
  pages = {7534--7550},
  year = {2020},
  address = {Online},
  publisher = {Association for Computational Linguistics},
  doi = {10.18653/v1/2020.emnlp-main.609},
  url = {https://aclanthology.org/2020.emnlp-main.609/}
}

@inproceedings{zhu2021tatqa,
  title = {{TAT-QA}: A Question Answering Benchmark on a Hybrid of Tabular and Textual Content in Finance},
  author = {Zhu, Fengbin and Lei, Wenqiang and Huang, Youcheng and Wang, Chao and Zhang, Shuo and Lv, Jiancheng and Feng, Fuli and Chua, Tat-Seng},
  booktitle = {Proceedings of the 59th Annual Meeting of the Association for Computational Linguistics and the 11th International Joint Conference on Natural Language Processing (Volume 1: Long Papers)},
  pages = {3277--3287},
  year = {2021},
  address = {Online},
  publisher = {Association for Computational Linguistics},
  doi = {10.18653/v1/2021.acl-long.254},
  url = {https://aclanthology.org/2021.acl-long.254/}
}

@inproceedings{min2023factscore,
  title = {{FActScore}: Fine-grained Atomic Evaluation of Factual Precision in Long Form Text Generation},
  author = {Min, Sewon and Krishna, Kalpesh and Lyu, Xinxi and Lewis, Mike and Yih, Wen-tau and Koh, Pang and Iyyer, Mohit and Zettlemoyer, Luke and Hajishirzi, Hannaneh},
  booktitle = {Proceedings of the 2023 Conference on Empirical Methods in Natural Language Processing},
  pages = {12076--12100},
  year = {2023},
  address = {Singapore},
  publisher = {Association for Computational Linguistics},
  doi = {10.18653/v1/2023.emnlp-main.741},
  url = {https://aclanthology.org/2023.emnlp-main.741/}
}

@inproceedings{manakul2023selfcheckgpt,
  title = {{SelfCheckGPT}: Zero-Resource Black-Box Hallucination Detection for Generative Large Language Models},
  author = {Manakul, Potsawee and Liusie, Adian and Gales, Mark},
  booktitle = {Proceedings of the 2023 Conference on Empirical Methods in Natural Language Processing},
  pages = {9004--9017},
  year = {2023},
  address = {Singapore},
  publisher = {Association for Computational Linguistics},
  doi = {10.18653/v1/2023.emnlp-main.557},
  url = {https://aclanthology.org/2023.emnlp-main.557/}
}

@article{rashkin2023attribution,
  title = {Measuring Attribution in Natural Language Generation Models},
  author = {Rashkin, Hannah and Nikolaev, Vitaly and Lamm, Matthew and Aroyo, Lora and Collins, Michael and Das, Dipanjan and Petrov, Slav and Tomar, Gaurav Singh and Turc, Iulia and Reitter, David},
  journal = {Computational Linguistics},
  volume = {49},
  number = {4},
  pages = {777--840},
  year = {2023},
  address = {Cambridge, MA, USA},
  publisher = {MIT Press},
  doi = {10.1162/coli_a_00486},
  url = {https://aclanthology.org/2023.cl-4.2/}
}

@inproceedings{gao2023alce,
  title = {Enabling Large Language Models to Generate Text with Citations},
  author = {Gao, Tianyu and Yen, Howard and Yu, Jiatong and Chen, Danqi},
  booktitle = {Proceedings of the 2023 Conference on Empirical Methods in Natural Language Processing},
  pages = {6465--6488},
  year = {2023},
  address = {Singapore},
  publisher = {Association for Computational Linguistics},
  doi = {10.18653/v1/2023.emnlp-main.398},
  url = {https://aclanthology.org/2023.emnlp-main.398/}
}

@inproceedings{liu2023geval,
  title = {{G}-Eval: {NLG} Evaluation Using {GPT}-4 with Better Human Alignment},
  author = {Liu, Yang and Iter, Dan and Xu, Yichong and Wang, Shuohang and Xu, Ruochen and Zhu, Chenguang},
  booktitle = {Proceedings of the 2023 Conference on Empirical Methods in Natural Language Processing},
  pages = {2511--2522},
  year = {2023},
  address = {Singapore},
  publisher = {Association for Computational Linguistics},
  doi = {10.18653/v1/2023.emnlp-main.153},
  url = {https://aclanthology.org/2023.emnlp-main.153/}
}

@inproceedings{wilder2020complement,
  title = {Learning to Complement Humans},
  author = {Wilder, Bryan and Horvitz, Eric and Kamar, Ece},
  booktitle = {Proceedings of the Twenty-Ninth International Joint Conference on Artificial Intelligence},
  pages = {1526--1533},
  year = {2020},
  publisher = {International Joint Conferences on Artificial Intelligence Organization},
  doi = {10.24963/ijcai.2020/212},
  url = {https://doi.org/10.24963/ijcai.2020/212}
}

@inproceedings{de2021assistance,
  title = {Classification Under Human Assistance},
  author = {De, Abir and Okati, Nastaran and Zarezade, Ali and {Gomez Rodriguez}, Manuel},
  booktitle = {Proceedings of the AAAI Conference on Artificial Intelligence},
  volume = {35},
  number = {7},
  pages = {5905--5913},
  year = {2021},
  doi = {10.1609/aaai.v35i7.16738},
  url = {https://ojs.aaai.org/index.php/AAAI/article/view/16738}
}

@inproceedings{okati2021triage,
  title = {Differentiable Learning Under Triage},
  author = {Okati, Nastaran and De, Abir and Gomez-Rodriguez, Manuel},
  booktitle = {Advances in Neural Information Processing Systems},
  volume = {34},
  pages = {9140--9151},
  year = {2021},
  publisher = {Curran Associates, Inc.},
  url = {https://proceedings.neurips.cc/paper/2021/hash/4c4c937b67cc8d785cea1e42ccea185c-Abstract.html}
}

@inproceedings{verma2022calibrated,
  title = {Calibrated Learning to Defer with One-vs-All Classifiers},
  author = {Verma, Rajeev and Nalisnick, Eric},
  booktitle = {Proceedings of the 39th International Conference on Machine Learning},
  pages = {22184--22202},
  year = {2022},
  volume = {162},
  series = {Proceedings of Machine Learning Research},
  publisher = {PMLR},
  url = {https://proceedings.mlr.press/v162/verma22c.html}
}

@inproceedings{narasimhan2022posthoc,
  title = {Post-hoc Estimators for Learning to Defer to an Expert},
  author = {Narasimhan, Harikrishna and Jitkrittum, Wittawat and Menon, Aditya K. and Rawat, Ankit and Kumar, Sanjiv},
  booktitle = {Advances in Neural Information Processing Systems},
  volume = {35},
  pages = {29292--29304},
  year = {2022},
  publisher = {Curran Associates, Inc.},
  url = {https://proceedings.neurips.cc/paper_files/paper/2022/hash/bc8f76d9caadd48f77025b1c889d2e2d-Abstract-Conference.html}
}

@article{jiang2021lmcalibration,
  title = {How Can We Know When Language Models Know? On the Calibration of Language Models for Question Answering},
  author = {Jiang, Zhengbao and Araki, Jun and Ding, Haibo and Neubig, Graham},
  journal = {Transactions of the Association for Computational Linguistics},
  volume = {9},
  pages = {962--977},
  year = {2021},
  address = {Cambridge, MA, USA},
  publisher = {MIT Press},
  doi = {10.1162/tacl_a_00407},
  url = {https://aclanthology.org/2021.tacl-1.57/}
}

@article{lin2024confidence,
  title = {Generating with Confidence: Uncertainty Quantification for Black-box Large Language Models},
  author = {Lin, Zhen and Trivedi, Shubhendu and Sun, Jimeng},
  journal = {Transactions on Machine Learning Research},
  year = {2024},
  issn = {2835-8856},
  url = {https://openreview.net/forum?id=DWkJCSxKU5}
}

@article{farquhar2024semanticentropy,
  title = {Detecting Hallucinations in Large Language Models Using Semantic Entropy},
  author = {Farquhar, Sebastian and Kossen, Jannik and Kuhn, Lorenz and Gal, Yarin},
  journal = {Nature},
  volume = {630},
  pages = {625--630},
  year = {2024},
  doi = {10.1038/s41586-024-07421-0},
  url = {https://www.nature.com/articles/s41586-024-07421-0}
}

@article{ji2023hallucinationsurvey,
  title = {Survey of Hallucination in Natural Language Generation},
  author = {Ji, Ziwei and Lee, Nayeon and Frieske, Rita and Yu, Tiezheng and Su, Dan and Xu, Yan and Ishii, Etsuko and Bang, Ye Jin and Madotto, Andrea and Fung, Pascale},
  journal = {ACM Computing Surveys},
  volume = {55},
  number = {12},
  articleno = {248},
  pages = {1--38},
  year = {2023},
  publisher = {Association for Computing Machinery},
  doi = {10.1145/3571730},
  url = {https://doi.org/10.1145/3571730}
}

@inproceedings{xiong2024expressuncertainty,
  title = {Can {LLM}s Express Their Uncertainty? An Empirical Evaluation of Confidence Elicitation in {LLM}s},
  author = {Xiong, Miao and Hu, Zhiyuan and Lu, Xinyang and Li, Yifei and Fu, Jie and He, Junxian and Hooi, Bryan},
  booktitle = {The Twelfth International Conference on Learning Representations},
  year = {2024},
  url = {https://openreview.net/forum?id=gjeQKFxFpZ}
}

@inproceedings{fadeeva2024factchecking,
  title = {Fact-Checking the Output of Large Language Models via Token-Level Uncertainty Quantification},
  author = {Fadeeva, Ekaterina and Rubashevskii, Aleksandr and Shelmanov, Artem and Petrakov, Sergey and Li, Haonan and Mubarak, Hamdy and Tsymbalov, Evgenii and Kuzmin, Gleb and Panchenko, Alexander and Baldwin, Timothy and Nakov, Preslav and Panov, Maxim},
  booktitle = {Findings of the Association for Computational Linguistics: ACL 2024},
  pages = {9367--9385},
  year = {2024},
  address = {Bangkok, Thailand},
  publisher = {Association for Computational Linguistics},
  doi = {10.18653/v1/2024.findings-acl.558},
  url = {https://aclanthology.org/2024.findings-acl.558/}
}

@inproceedings{gao2023rarr,
  title = {{RARR}: Researching and Revising What Language Models Say, Using Language Models},
  author = {Gao, Luyu and Dai, Zhuyun and Pasupat, Panupong and Chen, Anthony and Chaganty, Arun Tejasvi and Fan, Yicheng and Zhao, Vincent and Lao, Ni and Lee, Hongrae and Juan, Da-Cheng and Guu, Kelvin},
  booktitle = {Proceedings of the 61st Annual Meeting of the Association for Computational Linguistics (Volume 1: Long Papers)},
  pages = {16477--16508},
  year = {2023},
  address = {Toronto, Canada},
  publisher = {Association for Computational Linguistics},
  doi = {10.18653/v1/2023.acl-long.910},
  url = {https://aclanthology.org/2023.acl-long.910/}
}

@inproceedings{gou2024critic,
  title = {{CRITIC}: Large Language Models Can Self-Correct with Tool-Interactive Critiquing},
  author = {Gou, Zhibin and Shao, Zhihong and Gong, Yeyun and Shen, Yelong and Yang, Yujiu and Duan, Nan and Chen, Weizhu},
  booktitle = {The Twelfth International Conference on Learning Representations},
  year = {2024},
  url = {https://openreview.net/forum?id=Sx038qxjek}
}

@inproceedings{li2025rac,
  title = {{RAC}: Efficient {LLM} Factuality Correction with Retrieval Augmentation},
  author = {Li, Changmao and Flanigan, Jeffrey},
  booktitle = {Findings of the Association for Computational Linguistics: EMNLP 2025},
  pages = {25145--25159},
  year = {2025},
  address = {Suzhou, China},
  publisher = {Association for Computational Linguistics},
  doi = {10.18653/v1/2025.findings-emnlp.1370},
  url = {https://aclanthology.org/2025.findings-emnlp.1370/}
}

@inproceedings{zheng2023judge,
  title = {Judging {LLM}-as-a-Judge with {MT}-Bench and Chatbot Arena},
  author = {Zheng, Lianmin and Chiang, Wei-Lin and Sheng, Ying and Zhuang, Siyuan and Wu, Zhanghao and Zhuang, Yonghao and Lin, Zi and Li, Zhuohan and Li, Dacheng and Xing, Eric P. and Zhang, Hao and Gonzalez, Joseph E. and Stoica, Ion},
  booktitle = {Advances in Neural Information Processing Systems},
  volume = {36},
  pages = {46595--46623},
  year = {2023},
  address = {New Orleans, LA, USA},
  publisher = {Curran Associates, Inc.},
  url = {https://proceedings.neurips.cc/paper_files/paper/2023/hash/91f18a1287b398d378ef22505bf41832-Abstract-Datasets_and_Benchmarks.html}
}

\clearpage
\appendix
\setcounter{table}{0}
\renewcommand{\thetable}{\Alph{table}}

\section{Mechanism Diagnostics}

This section reports the complete budget ladder and paired bootstrap uncertainty supporting the main allocation result.
Table~\ref{tab:full_ladder} keeps operational baselines, diagnostic controls, and oracle references separate so that similar WAER values are not mistaken for equivalent post-review outcomes.

\begin{table}
\centering
\caption{Complete deterministic ladder over budgets (32-seed mean).
Each cell reports WAER/PRRE; lower is better.}
\label{tab:full_ladder}
\begin{tabular}{@{}lcccc@{}}
\toprule
Policy & 5\% & 10\% & 20\% & 40\% \\
\midrule
Random & .950/.978 & .900/.956 & .800/.911 & .600/.823 \\
Self-conf. & .949/.951 & .900/.905 & .793/.823 & .590/.718 \\
Rule/surface & .944/.944 & .893/.893 & .794/.813 & .591/.615 \\
Explicit & .900/.900 & .800/.800 & .600/.600 & .438/.556 \\
Scope-aware & .900/.900 & .800/.833 & .600/.767 & .223/.556 \\
Same-risk $r_i c_i$ & .900/.900 & .800/.800 & .600/.600 & .389/.556 \\
Same-risk $r_i c_i h_i$ & .900/.900 & .800/.800 & .600/.667 & .200/.556 \\
Risk-only & .900/.989 & .801/.962 & .605/.881 & .274/.704 \\
Review-value & .900/.903 & .800/.828 & .600/.716 & .219/.556 \\
Repair-count $y_i\rho_i$ & .900/.900 & .800/.800 & .600/.600 & .434/.556 \\
Gold-factor RV & .900/.900 & .800/.811 & .600/.656 & .200/.556 \\
Gold $y_i h_i$ & .900/1.000 & .800/1.000 & .600/.977 & .200/.756 \\
\bottomrule
\end{tabular}
\end{table}

The full ladder exposes the mechanism most clearly at small budgets.
At 5\% capacity, risk-only and review-value ranking have the same WAER (.900), but their PRRE values differ sharply (.989 versus .903).
The Gold $y_i h_i$ reference multiplies gold wrongness by the hand-set impact prior; it further shows why detecting consequential wrong answers is not sufficient, since its WAER reaches .200 at 40\% capacity while its PRRE remains .756 because selection is not conditioned on deterministic recoverability.
The gold-factor RV row uses gold wrongness and error type in the benchmark scoring rule, whereas the repair-count oracle directly ranks $y_i\rho_i$ and therefore gives the PRRE lower bound under the count budget.
As capacity grows, review-value ranking reaches the repair-count lower bound of .556 at 40\%.

\begin{table}
\centering
\caption{Paired, dataset-stratified source-fact bootstrap of the 32-seed mean at 20\% budget (1{,}000 resamples).
Queues are reranked per resample; negative $\Delta$ favors review-value.}
\label{tab:source_cluster_bootstrap}
\begin{tabular}{@{}llc@{}}
\toprule
Metric & Estimate & Point [95\% interval] \\
\midrule
WAER & Risk-only & .605 [.604,.607] \\
 & Review-value & .600 [.600,.600] \\
 & $\Delta$ & -.005 [-.007,-.004] \\
\addlinespace
WDE & Risk-only & .549 [.547,.553] \\
 & Review-value & .668 [.664,.671] \\
 & $\Delta$ & .119 [.114,.122] \\
\addlinespace
RVE & Risk-only & .626 [.622,.630] \\
 & Review-value & .510 [.508,.511] \\
 & $\Delta$ & -.116 [-.120,-.112] \\
\addlinespace
PRRE & Risk-only & .881 [.876,.885] \\
 & Review-value & .716 [.713,.720] \\
 & $\Delta$ & -.165 [-.168,-.160] \\
\bottomrule
\end{tabular}
\end{table}

Table~\ref{tab:source_cluster_bootstrap} complements Table~\ref{tab:review_value}'s outer-seed intervals by paired resampling of 120 source-fact clusters within dataset.
Each resample retains matched variants, recomputes every seed-specific queue, and averages metrics over seeds.
The WAER interval excludes zero, but the .005 effect remains operationally small at the .600 floor; the PRRE and RVE intervals favor review-value ranking.
The positive WDE difference records the intended severity--repairability trade-off rather than a uniform improvement on every exposure metric.

\section{LLM-Assisted Pilot}

This section reports a 200-item bridge study in which an LLM-assisted verifier supplies the risk score while the controlled benchmark continues to provide gold wrongness and outcome-side repair rules.
The pilot is not a replacement for the deterministic main result; it tests whether the same evaluation framework remains informative with a stronger, model-derived ranking signal.
Tables~\ref{tab:llm_curve}--\ref{tab:llm_error_breakdown} report the budget curve, the same-risk queue-objective ablation, and the error-type breakdown.

\begin{table}
\centering
\caption{LLM-assisted verifier pilot over budgets on 200 public-data-derived items.
Each cell reports WAER/PRRE; lower is better.}
\label{tab:llm_curve}
\begin{tabular}{@{}lcccc@{}}
\toprule
Strategy & 5\% & 10\% & 20\% & 40\% \\
\midrule
Random & .949/.978 & .898/.953 & .803/.909 & .603/.816 \\
Self-conf. & .930/.950 & .890/.920 & .760/.830 & .540/.720 \\
Rule/surface & .950/.970 & .920/.940 & .810/.840 & .570/.600 \\
Explicit & .940/.940 & .890/.890 & .750/.750 & .450/.540 \\
Strong verifier & .900/.910 & .810/.850 & .640/.750 & .410/.660 \\
Rule $\rightarrow$ verifier & .900/.900 & .800/.820 & .670/.720 & .420/.630 \\
Gold $y_i h_i$ & .900/1.000 & .800/1.000 & .600/.980 & .200/.740 \\
\bottomrule
\end{tabular}
\end{table}

At 20\% capacity, multiplying the same LLM risk score by affordance lowers PRRE from .750 to .640 while leaving WAER nearly unchanged (.640 to .630).
Adding the impact prior instead produces the lowest WAER and WDE but a higher PRRE (.720), again showing that answer capture, severity reduction, and deterministic repair exposure are distinct objectives.

\begin{table}
\centering
\caption{LLM-assisted same-risk ablation at 20\% budget on the 200-item pilot; lower is better.}
\label{tab:llm_same_risk}
\begin{tabular}{lccc}
\toprule
Queue objective & WAER & PRRE & WDE \\
\midrule
LLM $r_i$ & .640 & .750 & .688 \\
$r_i c_i$ & .630 & \textbf{.640} & .774 \\
$r_i c_i h_i$ & \textbf{.600} & .720 & \textbf{.643} \\
Gold-factor RV & \textbf{.600} & \textbf{.640} & .719 \\
\bottomrule
\end{tabular}
\end{table}

\begin{table}
\centering
\caption{LLM-assisted strong-verifier error-type breakdown at 20\% budget; lower WAER is better.}
\label{tab:llm_error_breakdown}
\begin{tabular}{@{}lcc@{}}
\toprule
Error type & WAER & $n$ \\
\midrule
Conclusion mismatch & .697 & 33 \\
Direction flip & .824 & 17 \\
Numeric perturbation & .000 & 17 \\
Scope distortion & 1.000 & 17 \\
Unsupported addition & .625 & 16 \\
\bottomrule
\end{tabular}
\end{table}

The error-type breakdown is directionally consistent with the controlled mechanism result for the easiest and hardest intervention classes: all numeric perturbations are selected, whereas all scope distortions remain exposed.
Direction flips are harder for this verifier (WAER .824), which cautions against treating the pilot as a model-independent performance estimate.
Together, these results support the queue-objective diagnosis while preserving the pilot's auxiliary status.

\paragraph{Repair-side pilot checks.}
We do not use model-side repair checks as gold labels.
In a 100-answer pilot, benchmark-side success was .95 for direction flips and numeric perturbations, .73 for conclusion mismatches, and .30 for unsupported additions, reflecting conservative rules where correction can blur into abstention.
Separately, six independent judge models rated 251/300 repairs (83.7\%) as reducing wrong-answer exposure and 240/300 (80.0\%) as supported by public evidence, with no parse failures.
The panel comprised Qwen3.6 Flash, Gemini 2.5 Flash Lite, GPT-4o-mini, Claude 3.5 Haiku, GPT-5.5, and Mistral Medium 3.5.
Agreement was lowest for numeric perturbations (38/60), likely because several judges treated arithmetic-expression evidence conservatively.
These diagnostic checks do not affect WAER or PRRE.

\section{Base-Rate and Prior Sensitivity}
\label{app:base_rate}

This section reports robustness checks for two assumptions of the balanced stress benchmark: the prevalence of wrong answers and the hand-set affordance priors.
Table~\ref{tab:base_rate} changes the wrong-answer rate while holding the 20\% count budget fixed.
At the sparsest 5\% rate, WAER is effectively tied (.155 versus .156) while review-value ranking still lowers PRRE (.632 to .551); from 10\% through 30\%, it lowers both metrics.
This pattern separates count-budget saturation from the repair-aware allocation effect.

\begin{table}
\centering
\caption{Wrong-rate sweep at 20\% budget, 32-seed mean; lower is better.}
\label{tab:base_rate}
\begin{tabular}{lcccc}
\toprule
 & \multicolumn{2}{c}{WAER} & \multicolumn{2}{c}{PRRE} \\
\cmidrule(lr){2-3}\cmidrule(lr){4-5}
Wrong rate & risk & review & risk & review \\
\midrule
5\% & .155 & .156 & .632 & .551 \\
10\% & .184 & .168 & .651 & .541 \\
15\% & .243 & .195 & .681 & .555 \\
20\% & .280 & .212 & .707 & .547 \\
25\% & .327 & .229 & .737 & .562 \\
30\% & .401 & .331 & .778 & .595 \\
\bottomrule
\end{tabular}
\end{table}

\begin{table}
\centering
\caption{Prior sensitivity at 20\% budget; lower PRRE is better.}
\label{tab:sensitivity}
\begin{tabular}{lcc}
\toprule
$\delta$ & risk-only PRRE & review-value PRRE \\
\midrule
-0.2 & .881 & \textbf{.604} \\
-0.1 & .881 & \textbf{.652} \\
0.0 & .881 & \textbf{.716} \\
0.1 & .881 & \textbf{.802} \\
0.2 & \textbf{.881} & .888 \\
\bottomrule
\end{tabular}
\end{table}

\begin{table}
\centering
\caption{Random prior sensitivity at 20\% budget.
Deltas are review-value minus risk-only; negative favors review-value.}
\label{tab:random_prior_sensitivity}
\begin{tabular}{@{}llcc@{}}
\toprule
Radius & Metric & Win & $\Delta$ [range] \\
\midrule
$\pm .1$ & RVE & 100.0\% & -.125 [-.238,-.059] \\
 & PRRE & 100.0\% & -.161 [-.262,-.045] \\
\addlinespace
$\pm .2$ & RVE & 100.0\% & -.151 [-.331,-.054] \\
 & PRRE & 94.3\% & -.152 [-.281,.017] \\
\addlinespace
$\pm .3$ & RVE & 100.0\% & -.192 [-.431,-.049] \\
 & PRRE & 84.3\% & -.149 [-.281,.048] \\
\bottomrule
\end{tabular}
\end{table}

\begin{table}
\centering
\caption{Template-cue and cost-proxy sanity checks at 20\% budget.
Paraphrase rows report PRRE after deterministic surface rewrites.
Cost rows compare count-budget and surface-cost-budget selection.
Lower PRRE is better.}
\label{tab:template_cost_sanity}
\begin{tabular}{lccc}
\toprule
Check & Risk PRRE & Review PRRE & $\Delta$ \\
\midrule
Original templates & .881 & \textbf{.716} & -.165 \\
Light paraphrase & .852 & \textbf{.716} & -.136 \\
Cue-stripped paraphrase & .798 & \textbf{.721} & -.077 \\
\midrule
Count budget & .881 & \textbf{.716} & -.165 \\
Surface-cost budget & .877 & \textbf{.722} & -.155 \\
\bottomrule
\end{tabular}
\end{table}

The prior checks make the limitations of the hand-set affordance priors explicit.
Under a fixed positive shift of .2, review-value PRRE becomes slightly worse than risk-only (.888 versus .881), so the advantage is not invariant to arbitrary priors.
Across random perturbations, however, PRRE still favors review value in 100.0\%, 94.3\%, and 84.3\% of trials at radii .1, .2, and .3, respectively, while RVE favors it in every trial.
The template and surface-cost checks retain the qualitative PRRE advantage, suggesting that neither simple lexical cues nor a single count-cost assumption explains the main result.

\section{Reproducibility Notes}

The deterministic lane contains 720 instances in 120 source clusters: 60 TAT-QA facts with eight variants each and 60 SciFact facts with four variants each.
Runs use 32 seeds, four budgets (5--40\%), 5--30\% wrong-rate sweeps, paraphrase/cost checks, and 1{,}000 paired, dataset-stratified source-cluster resamples.
Each resample retains matched variants, reranks queues, and averages over seeds.
Gold labels derive from public labels and controlled mutations.
Repairable wrong-answer counts are 152 at the base seed and average 159.7 because unsupported-addition templates vary.
Stable ties preserve seeded order; \texttt{scripts/pilot\_waer\_stress.py} records the formulas, rules, and priors.
Model rows record provider/model snapshots (default \texttt{openai/gpt-5.5}; $T=.2$ for paraphrasing, 0 otherwise).

\end{document}